\documentclass{article}
\usepackage{spconf,amsmath,amssymb,graphicx}
\usepackage[T1]{fontenc}
\usepackage[utf8]{inputenc}
\usepackage{booktabs,array,tabularx,multirow}
\usepackage{needspace}
\usepackage[table]{xcolor}
\usepackage[hidelinks]{hyperref}
\definecolor{tableblue}{RGB}{235,242,247}
\definecolor{tablegreen}{RGB}{232,242,238}
\title{EviSI: An Evidence-Based Evaluation Agent\\for Simultaneous Interpreting}
\name{\begin{tabular}{@{}c@{}}
Ben Yan$^{1,*}$, Zongyao Li$^{1,*}$, Xiaoyu Chen$^{1}$, Daimeng Wei$^{1,\dagger}$,\\
Weidong Liu$^{1}$, Huan Zhao$^{1}$, Chong Li$^{1}$, Yaode Wang$^{1}$, Yuzhe Shang$^{2}$
\end{tabular}\thanks{{\small$^{*}$Equal contribution. $^{\dagger}$Corresponding author.\\Our code is available at \href{https://github.com/caiqiezujian/EviSI-Eval}{https://github.com/caiqiezujian/EviSI-Eval}.}}}
\address{$^{1}$Huawei Translation Service Center, Beijing, China\\
$^{2}$School of Informatics, Xiamen University, China\\
{\normalsize\{yanben4, lizongyao\}@huawei.com; chenxiaoyu35@h-partners.com}\\
{\normalsize\{weidaimeng, oliver.liuweidong, zhaohuan54\}@huawei.com}\\
{\normalsize\{august.li, wangyaode1, shangyuzhe1\}@huawei.com}}
\hypersetup{
pdftitle={EviSI: An Evidence-Based Evaluation Agent for Simultaneous Interpreting},
pdfauthor={Ben Yan, Zongyao Li, Xiaoyu Chen, Daimeng Wei, Weidong Liu, Huan Zhao, Chong Li, Yaode Wang, Yuzhe Shang},
pdfkeywords={Speech-to-speech translation, simultaneous interpreting, quality evaluation, LLM agents}
}
\begin{document}
\maketitle
\begin{abstract}
\begingroup
\emergencystretch=2em
Low-latency simultaneous speech-to-speech translation must keep pace with ongoing speech while preserving key information. To meet these demands, systems use segmentation, reformulation and condensation to reorganize and rephrase information. However, metrics developed for text translation, including BLEU and COMET, may not consistently distinguish faithful adaptations from semantic errors. We propose EviSI, a large language model evaluation agent combining Multidimensional Quality Metrics (MQM) with criteria developed with professional interpreters. Shared source evidence guides assessment across four dimensions: Anchor, Event, Logic and Fluency. Verified errors are deduplicated before deterministic scoring. On human-rated English to Chinese and Chinese to English data, EviSI recovers the aggregate English to Chinese human system ranking. Mean within-dataset Kendall correlations for system rankings reach 0.707 and 0.467, respectively, exceeding evaluated BLEU and COMET baselines. A multilingual extension to five directions without human ratings retains the dimensions and scoring rule, showing positive system ranking correlations with COMET throughout.
\par
\endgroup
\end{abstract}
\begin{keywords}
Speech-to-speech translation, simultaneous interpreting, quality evaluation, LLM agents
\end{keywords}

\begin{figure}[t]
\centering
\includegraphics[width=\columnwidth]{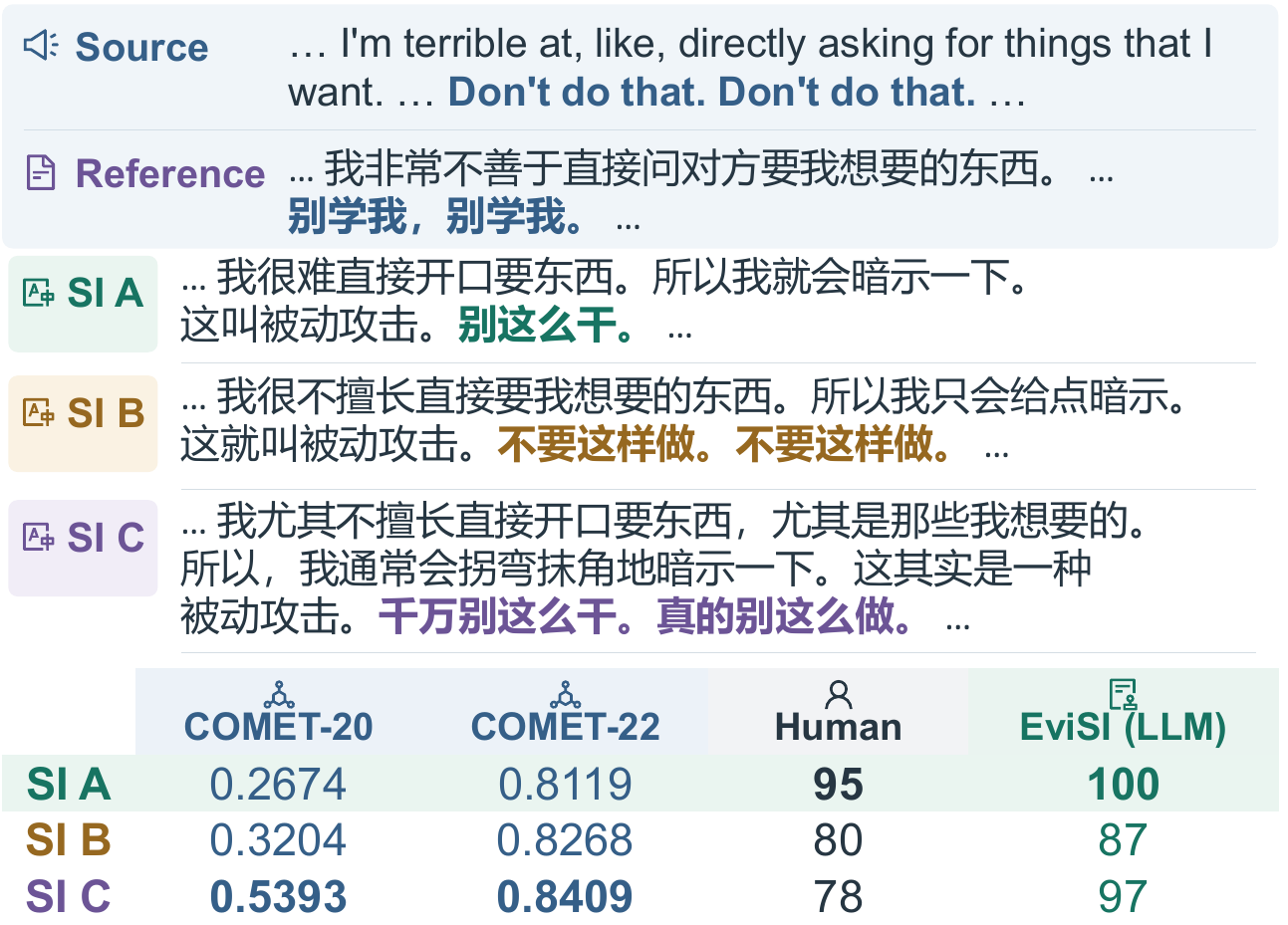}
\caption{Human evaluation and EviSI favor A, whereas COMET20/22 favor C. Scores cover complete outputs on original scales; text shows excerpts. Bold marks column maxima.}
\label{fig:case}
\end{figure}

\begin{figure*}[t]
\centering
\includegraphics[width=\textwidth]{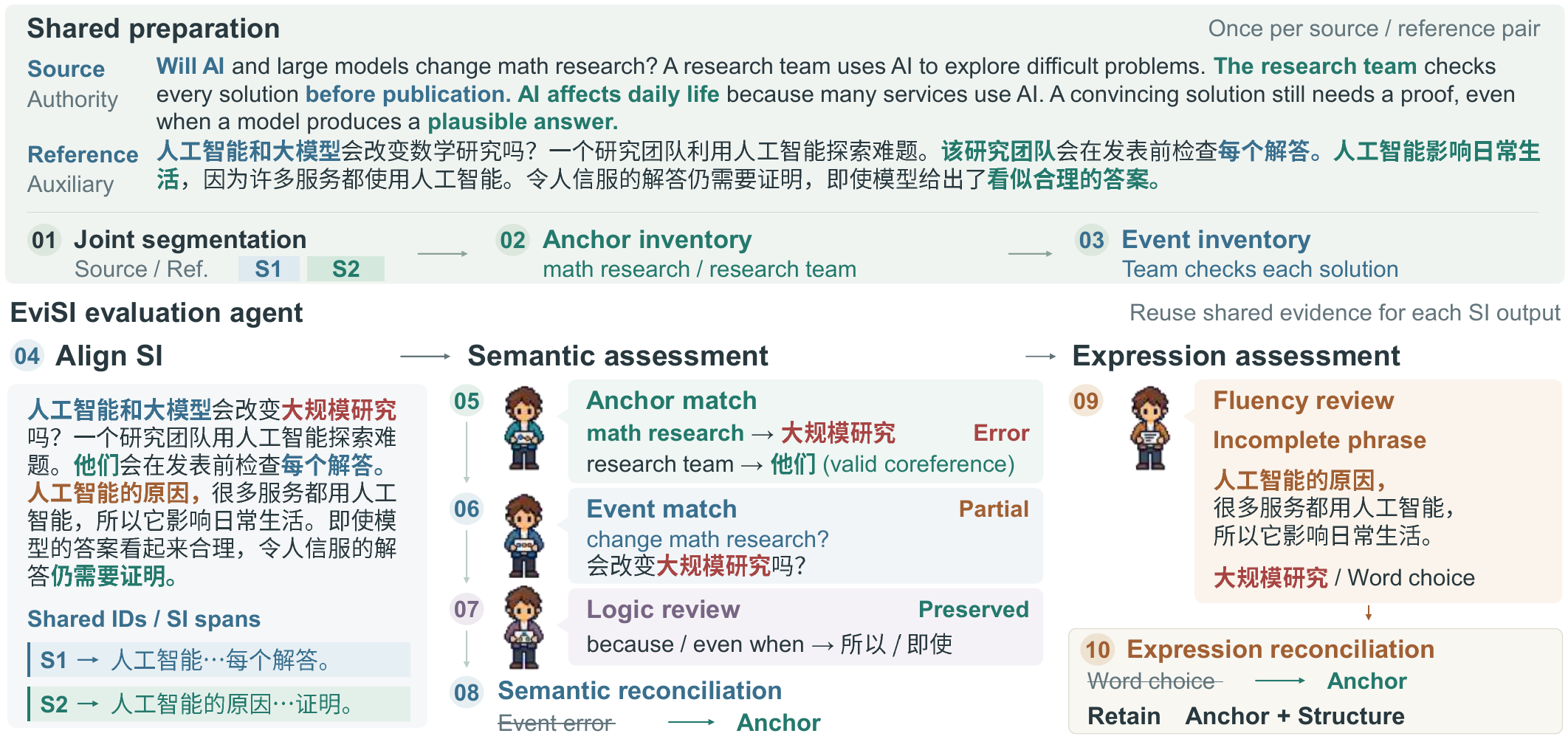}
\caption{EviSI with an adapted example. Shared preparation (01 to 03) precedes candidate assessment (04 to 10). Valid coreference is accepted; deduplication preserves distinct Anchor and Fluency errors.}
\label{fig:method}
\end{figure*}

\begin{table*}[t]
\centering
\caption{Agreement with human judgments: aggregate system ranks, mean corpus ranking agreement and individual output correlation. Bold marks column maxima, including ties, not significance.}
\label{tab:alignment}
\setlength{\tabcolsep}{4pt}
\renewcommand{\arraystretch}{1.08}
\begin{tabularx}{\textwidth}{l*{8}{>{\centering\arraybackslash}X}}
\toprule
& \multicolumn{4}{c}{EN$\rightarrow$ZH} & \multicolumn{4}{c}{ZH$\rightarrow$EN} \\
\cmidrule(lr){2-5}\cmidrule(lr){6-9}
\rowcolor{tableblue}
Method & Sys. $\rho$ & Sys. $\tau_b$ & Corpus $\bar\tau_b$ & Sample $r$ & Sys. $\rho$ & Sys. $\tau_b$ & Corpus $\bar\tau_b$ & Sample $r$ \\
\midrule
COMET20 & 0.314 & 0.200 & 0.387 & 0.243 & 0.314 & 0.200 & 0.267 & 0.402 \\
COMET22 & 0.314 & 0.200 & 0.280 & 0.208 & 0.143 & 0.067 & 0.333 & 0.404 \\
COMETKiwi & 0.314 & 0.200 & 0.413 & 0.377 & 0.143 & 0.067 & 0.333 & \textbf{0.442} \\
Sentence BLEU & 0.771 & 0.600 & 0.360 & 0.152 & 0.543 & \textbf{0.467} & 0.267 & 0.229 \\
\rowcolor{tablegreen}
\textbf{EviSI} & \textbf{1.000} & \textbf{1.000} & \textbf{0.707} & \textbf{0.436} & \textbf{0.600} & \textbf{0.467} & \textbf{0.467} & 0.432 \\
\bottomrule
\end{tabularx}
\end{table*}

\section{Introduction}
\label{sec:intro}
Simultaneous speech-to-speech translation (S2S) must understand unfolding speech and decide when to begin delivery~\cite{zhang2024streamspeech}. Unlike offline text translation, it faces incomplete context and speech understanding errors, whether from automatic speech recognition (ASR) in a cascade or direct translation~\cite{cheng2024clasi,zhang2024streamspeech}. Waiting can clarify meaning, but also increases listener lag~\cite{cheng2024clasi}.

Interpreters use segmentation and reformulation~\cite{he2016interpretese}. Simultaneous interpreting (SI) research also explores translating complete semantic chunks~\cite{cheng2024clasi} and condensing redundant expressions~\cite{zhang2026redefining}. These strategies aim to convey established meaning without unsupported guesses and shorten delivery while preserving key facts. Evaluation should allow faithful condensation while checking factual accuracy, logical consistency~\cite{zwischenberger2010quality} and clarity~\cite{kurz2001quality}.

BLEU~\cite{papineni2002bleu} and COMET~\cite{rei2020comet} remain useful, but may not consistently account for SI strategies~\cite{wein2024barriers}. In Fig.~\ref{fig:case}, A condenses a repeated warning without losing its core message, whereas C is more elaborate. For complete outputs, human evaluation and EviSI rank A first; COMET20/22 favor C. This disagreement motivates distinguishing acceptable condensation from semantic loss.

SI studies address word order~\cite{doi2026simul}, rating agreement~\cite{fantinuoli2024} and transcript alignment~\cite{xue2026practical}. MQM organizes evaluation by error type and severity~\cite{lommel2014mqm,freitag2021mqm}. Automated methods include learned error detection~\cite{perrella2022matese,guerreiro2024xcomet}, large language model (LLM) scoring~\cite{kocmi2023gemba} and error analysis~\cite{fernandes2023automqm,kocmi2023mqm}, refined through aggregation, rubrics and debate~\cite{junczys2025gemba,kim2025rubric,feng2025mmad}. Yet LLM judges may underuse source information~\cite{huang2024source}, motivating explicit checks against source evidence.

We propose EviSI (Evidence-based Evaluation for Simultaneous Interpreting), an LLM agent combining MQM with criteria developed with professional interpreters. Shared source evidence is fixed before candidate assessment across four dimensions: Anchor, Event, Logic and Fluency. Only verified, deduplicated errors incur penalties.

Across nine English and Chinese corpora, EviSI improves average human ranking agreement, with mixed individual score correlations. We also evaluate five directions, including English to Japanese, German and French, adapting language instructions while retaining the dimensions and penalties. System rankings correlate positively with COMET but lack human validation.

\begin{table*}[t]
\centering
\caption{Human / EviSI mean scores by corpus and system, on uncalibrated scales. Bold marks each evaluator's row maxima, including ties.}
\label{tab:scores}
\setlength{\tabcolsep}{4pt}
\renewcommand{\arraystretch}{1.08}
\begin{tabularx}{\textwidth}{l*{6}{>{\centering\arraybackslash}X}}
\toprule
\rowcolor{tableblue}
Corpus & H1 & H2 & H3 & H4 & H5 & H6 \\
\midrule
\multicolumn{7}{l}{\textit{English to Chinese}} \\
E1 & \textbf{78.20} / 76.37 & 75.30 / \textbf{77.88} & 75.70 / 73.95 & 73.85 / 71.18 & 73.88 / 71.81 & 74.78 / 74.75 \\
E2 & \textbf{82.58} / \textbf{76.53} & 75.63 / 67.26 & 74.84 / 72.88 & 76.32 / 69.92 & 77.34 / 73.32 & 75.82 / 69.93 \\
E3 & \textbf{77.70} / 78.43 & 77.35 / \textbf{80.16} & 76.23 / 77.73 & 71.25 / 76.41 & 69.98 / 73.82 & 73.73 / 76.98 \\
E4 & 61.89 / 72.52 & \textbf{68.25} / \textbf{78.38} & 62.19 / 73.26 & 63.14 / 75.31 & 60.39 / 71.85 & 58.58 / 67.28 \\
E5 & 78.43 / 71.60 & 84.60 / \textbf{78.62} & \textbf{85.95} / 76.74 & 82.18 / 78.15 & 80.85 / 75.22 & 76.10 / 73.49 \\
\midrule
\multicolumn{7}{l}{\textit{Chinese to English}} \\
Z1 & \textbf{80.75} / \textbf{80.15} & 78.75 / 78.05 & 78.64 / 75.53 & 77.32 / 75.11 & 79.79 / 76.80 & 76.82 / 74.54 \\
Z2 & \textbf{74.18} / \textbf{70.92} & 65.59 / 63.29 & 69.33 / 63.12 & 67.72 / 67.11 & 68.85 / 60.47 & 68.64 / 63.28 \\
Z3 & \textbf{76.13} / 77.82 & 74.32 / \textbf{81.68} & 70.82 / 77.38 & 71.71 / 80.42 & 72.05 / 81.23 & 70.05 / 77.68 \\
Z4 & \textbf{72.45} / 74.51 & 71.58 / \textbf{77.94} & 70.13 / 71.07 & 71.15 / 74.01 & 70.60 / 73.80 & 71.43 / 72.41 \\
\bottomrule
\end{tabularx}
\end{table*}

\section{Method}
\label{sec:method}
The LLM assesses meaning and expression through specialized roles; code validates report structure and computes scores (Fig.~\ref{fig:method}).

\subsection{Shared preparation before evaluation}
Inputs are complete source and SI texts, plus an auxiliary reference. Evaluation follows completion of the actual SI output. The source determines correctness; the reference helps locate and align evidence without prescribing wording.

Shared preparation segments source and reference without discarding text (01), then extracts \emph{Anchors} (entities, terms, quantities and times; 02) and \emph{Events} (statements, participants and conditions; 03). Inventories, evidence spans and importance labels are fixed per source and reference pair and reused across systems.

SI alignment (04) partitions each output to match shared segments without judging quality. Matching checks whether inventoried information is conveyed, accepting faithful paraphrase and clear pronoun references. This common checklist constrains evaluation; extraction and matching still require LLM judgments.

\subsection{Assessment and reconciliation}
\textbf{Semantic assessment (05 to 08).} Anchor checks key facts; Event checks statements and conditions; Logic checks participant roles, truth conditions, relations and modifiers. In Fig.~\ref{fig:method}, the pronoun correctly identifies the research team, but \emph{math research} is mistranslated. After Logic consolidation (07), semantic reconciliation (08) removes the Event report duplicating this Anchor error. Equivalent meanings and valid alternatives incur no deduction.

\textbf{Expression assessment (09 to 10).} Fluency checks word choice, structure, fillers, incomplete expressions and punctuation. Its reports (09) are compared with semantic errors (10). The example's word choice report duplicates the Anchor error; only the separate structural error adds a penalty. Text cannot assess pronunciation, voice quality or prosody.

Only reports describing the same error are merged; overlapping spans alone do not justify merging. All reports remain inspectable.

\Needspace{4\baselineskip}
\subsection{Deterministic scoring}
EviSI follows MQM's penalty principle~\cite{lommel2024mqm} but uses SI deductions without length normalization, rather than a standard MQM score. Both experiments use:
\begin{equation}
S=\max(0,100-D_A-D_E-D_L-D_F).
\label{eq:score}
\end{equation}
For $d\in\{A,E\}$, $D_d=\alpha_d\sum_{j\in\mathcal D_d^+}w_jc(v_j)$, where $\mathcal D_d^+$ contains retained errors, $w_j\in\{1,2,3\}$ denotes importance, and $(\alpha_A,\alpha_E)=(4,5)$. Verdict losses are 0 for equivalent or valid alternative, 0.5 for partial, 0.8 for missing and 1 for contradiction. Logic penalties follow the category order above: $(8,8,6,5)$; Fluency uses $(3,5,4,6,2)$. A combined module label receives only its largest component penalty. Rescoring existing reports assigns importance 1 when absent or zero, and loss 0.5 to unrecognized verdicts; module labels follow the original parsing rules. Fixed judgments produce identical scores, although repeated LLM calls may change the judgments.

\section{Experiments}
\label{sec:experiments}
\subsection{Data and evaluation protocol}
\textbf{Data and human evaluation.} Internal recordings of authentic, information-dense speech form five EN$\rightarrow$ZH corpora (E1 to E5) and four ZH$\rightarrow$EN corpora (Z1 to Z4). Six commercial S2S systems (H1 to H6) have 1,164 and 870 commonly scored outputs from 194 and 145 sources, respectively, without imputation. Multiple experienced professional interpreters used common MQM and SI criteria, prioritizing essential meaning, logical coherence and clear expression while allowing faithful reformulation, coreference and condensation. This retrospective analysis has no independent test set.

\textbf{Evaluator.} DeepSeek-V4-Pro judges errors and reconciles reports. Code computes scores from retained errors, rather than requesting LLM scores.

\textbf{Metrics.} Baselines on the same records are precomputed Sentence BLEU, COMET20~\cite{rei2020comet}, COMET22~\cite{rei2022comet22} and reference-free COMETKiwi~\cite{rei2022kiwi}. We average sentence BLEU scores rather than compute corpus BLEU. System comparison is the primary objective~\cite{kocmi2021ship}; ranking agreement and score correlation are evaluated separately~\cite{lavie2025metrics}. Spearman $\rho$ and Kendall $\tau_b$ compare six system means per direction, weighting corpora by source count. We also average corpus $\tau_b$ values equally. Pearson $r$ compares individual output scores. Paired bootstrap intervals resample sources within corpora, retaining all six outputs per source.

\subsection{Agreement with human system rankings}
EviSI matches all 15 pairwise human preferences in the aggregate EN$\rightarrow$ZH ranking (Table~\ref{tab:alignment}). Its $\tau_b=1.000$ exceeds Sentence BLEU's 0.600 and the COMET family's 0.200. For ZH$\rightarrow$EN, EviSI has the highest $\rho$ (0.600), matching 11 of 15 pairs; its $\tau_b=0.467$ ties Sentence BLEU. Table~\ref{tab:scores} reports human and EviSI means on their original scales, without calibration.

\begin{figure}[t]
\centering
\includegraphics[width=\columnwidth]{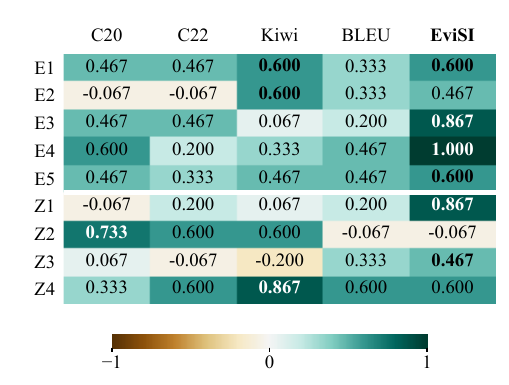}
\caption{Human system ranking agreement ($\tau_b$). Bold marks row maxima. C20/C22: COMET20/22; Kiwi: COMETKiwi; BLEU: Sentence BLEU.}
\label{fig:heatmap}
\end{figure}

With equal corpus weighting, mean system ranking correlations are 0.707 and 0.467 for EviSI, versus the strongest baselines' 0.413 and 0.333 (Fig.~\ref{fig:heatmap}). Across nine corpora, EviSI reaches $\bar\tau_b=0.600$, versus COMETKiwi's 0.378. On Z2, however, EviSI obtains $-0.067$ and COMET20 reaches 0.733. EviSI's average advantage therefore does not hold for every corpus.

For individual outputs, EviSI and COMETKiwi obtain $r=0.436$ and 0.377 in EN$\rightarrow$ZH, versus 0.432 and 0.442 in ZH$\rightarrow$EN. Paired bootstrap 95\% intervals for their differences (EviSI minus COMETKiwi) are $[-0.035,0.146]$ and $[-0.099,0.090]$. Both include zero, providing no clear evidence of improved correlation for individual output scores.

\subsection{Multilingual extension without human ratings}
Four commercial S2S systems (M1 to M4), distinct from H1 to H6, are evaluated on RealSI~\cite{cheng2024clasi} subsets for EN$\leftrightarrow$ZH and ACL 60/60~\cite{salesky2023acl6060} for EN$\rightarrow$JA/DE/FR. Language instructions are adapted, retaining the dimensions and penalties, with evidence prepared per source and reference pair.

Table~\ref{tab:multi} reports corpus means, equally averaged per direction on each metric's original scale. EviSI uses Eq.~\ref{eq:score}. Compare systems within each metric. Individual COMET inputs were not audited; other run settings are not verified across languages.

\begin{table}[t]
\centering
\small
\caption{Multilingual system means ($\uparrow$), without human ratings. Bold: highest within each direction and metric.}
\label{tab:multi}
\setlength{\tabcolsep}{2pt}
\renewcommand{\arraystretch}{0.92}
\setlength{\aboverulesep}{1pt}
\setlength{\belowrulesep}{1pt}
\begin{tabularx}{\columnwidth}{lc>{\hsize=1.04\hsize\linewidth=\hsize\centering\arraybackslash}X>{\hsize=1.04\hsize\linewidth=\hsize\centering\arraybackslash}X>{\hsize=1.26\hsize\linewidth=\hsize\centering\arraybackslash}X>{\hsize=.66\hsize\linewidth=\hsize\centering\arraybackslash}X}
\toprule
\rowcolor{tableblue}
Direction & System & COMET20 & COMET22 & COMETKiwi & EviSI \\
\midrule
 & M1 & 0.5184 & 0.8208 & \textbf{0.6990} & 59.67 \\
 & M2 & 0.4008 & 0.8001 & 0.6809 & 56.70 \\
 & M3 & \textbf{0.5859} & \textbf{0.8401} & 0.6987 & \textbf{64.96} \\
\multirow{-4}{*}{\textbf{EN$\rightarrow$ZH}} & M4 & 0.1718 & 0.7580 & 0.6712 & 35.60 \\
\arrayrulecolor{black!35}
\midrule
\arrayrulecolor{black}
\rowcolor{black!4}
 & M1 & 0.4360 & 0.7924 & 0.6520 & 55.00 \\
\rowcolor{black!4}
 & M2 & 0.3040 & 0.7689 & 0.6460 & 49.15 \\
\rowcolor{black!4}
 & M3 & \textbf{0.5496} & \textbf{0.8037} & \textbf{0.6613} & \textbf{65.17} \\
\rowcolor{black!4}
\multirow{-4}{*}{\textbf{ZH$\rightarrow$EN}} & M4 & 0.0367 & 0.7364 & 0.6390 & 36.34 \\
\arrayrulecolor{black!35}
\midrule
\arrayrulecolor{black}
 & M1 & 0.5524 & 0.8641 & 0.7639 & 38.37 \\
 & M2 & 0.5297 & 0.8644 & \textbf{0.7640} & \textbf{60.21} \\
 & M3 & \textbf{0.6566} & \textbf{0.8787} & 0.7562 & 47.31 \\
\multirow{-4}{*}{\textbf{EN$\rightarrow$JA}} & M4 & 0.3652 & 0.8120 & 0.6872 & 23.34 \\
\arrayrulecolor{black!35}
\midrule
\arrayrulecolor{black}
\rowcolor{black!4}
 & M1 & \textbf{0.5747} & \textbf{0.7880} & \textbf{0.7342} & \textbf{63.44} \\
\rowcolor{black!4}
 & M2 & 0.4465 & 0.7668 & 0.7217 & 59.76 \\
\rowcolor{black!4}
 & M3 & 0.5022 & 0.7741 & 0.7241 & 56.01 \\
\rowcolor{black!4}
\multirow{-4}{*}{\textbf{EN$\rightarrow$FR}} & M4 & 0.1463 & 0.6984 & 0.6299 & 35.78 \\
\arrayrulecolor{black!35}
\midrule
\arrayrulecolor{black}
 & M1 & \textbf{0.5296} & \textbf{0.7904} & \textbf{0.7182} & \textbf{51.58} \\
 & M2 & 0.5257 & 0.7900 & 0.7106 & 47.40 \\
 & M3 & 0.4687 & 0.7768 & 0.6910 & 47.76 \\
\multirow{-4}{*}{\textbf{EN$\rightarrow$DE}} & M4 & 0.1639 & 0.6999 & 0.6143 & 25.23 \\
\bottomrule
\end{tabularx}
\end{table}

All evaluators rank M4 last throughout. EviSI and COMET20/22 rank M3 first for EN$\leftrightarrow$ZH, but M2 and M3, respectively, for EN$\rightarrow$JA. For French and German, all rank M1 first, differing only in M2/M3 ordering. Mean Kendall $\tau_b$ with COMET20, COMET22 and COMETKiwi is 0.733, 0.800 and 0.733. Human ratings are needed to determine which ordering better reflects quality.

\section{Conclusion}
EviSI combines SI criteria derived from MQM with shared evidence to assess semantic preservation and expression. Experiments show improved average agreement with human system rankings. Across five directions without human ratings, system rankings correlate positively with COMET.

Deterministic scoring does not eliminate uncertainty in LLM judgments. Future work will strengthen evidence verification to better constrain semantic judgments, conduct human evaluation across languages, and assess latency, pronunciation and prosody in real interpreting settings. These extensions will also draw on listener studies of SI~\cite{javorsky2022}.
\label{endtechnical}

\clearpage
\begingroup
\ninept
\bibliographystyle{IEEEbib}
\bibliography{references}
\endgroup
\end{document}